# Predict Future Sales using Ensembled Random Forests

**Yuwei Zhang**
Peking University
*1600012935@pku.edu.cn*

**Xin Wu**
Peking University
*1600013025@pku.edu.cn*

**Chengyang Gu**
Peking University
*1600015431@pku.edu.cn*

**Yueqi Xie**
Peking University
*1600012724@pku.edu.cn*

## Abstract

In this paper, we propose a rather simple approach to future sales predicting based on feature engineering, Random Forest Regressor and ensemble learning. Its performance turned out to exceed many of the conventional methods and get final score 0.88186, representing root mean squared error. As of this writing, our model ranked 5th on the leaderboard.

## 1     Introduction

Predict Future Sales is a kaggle competition relating to time series prediction. In this competition the participants work with a challenging time-series dataset consisting of daily sales data, kindly provided by one of the largest Russian software firms - 1C Company. We are asked to predict total sales for every product and store in the next month.

The main steps of the predicting include feature engineering and regressing analysis. Better feature engineering method significantly give rise to estimation accuracy. Experimentally, we found that the feature engineering methods proposed by Denis Larionov in *Feature engineering, xgboost* can be an excellent preprocessing method. Thus we pay more attention on choosing a suitable regressor for this specified application. Conventional methods involve LSTM, XGBOOST and LightGBM, which are commonly used in time series predicting. But we find that to achieve further achievement, simple regressor like random forest can keep the characteristic of the features and gain a better result. We also tried ensemble learning to further enhance our model.

## 2     Related Works

There are several approaches introduced in kernels with relatively high accuracy, and we introduce some of those models we referred to in the following part, emphasizing on their main ideas and performance.

*Nested LSTM*: This model is proposed by James Lee in the kernel *Predicting sales with a nested LSTM*. This model is based on Keras implementation of Nested LSTMs with two layers. The way that the training set is built is to convert the raw sales data to monthly sales, broken out by item and shop. The Root Mean Squared Error of this method achieved 1.02.

*Feature engineering, xgboost*: This model is proposed by Denis Larionov in the kernel *Feature engineering, xgboost*. The model is based on XGBRegressor and feature engineering including the removal of outliers and several special process of features. The Root Mean Squared Error of this method achieved 0.91.

# 3 Methodology

Our aim is to make an as accurate as possible a priori prediction about total sales for any given product and store that sells it, provided with historical sales data, using the method of machine learning algorithms. In theory this involves dealing with a time series dataset.

## 3.1 Preprocessing

To construct a model that may effectively perform the operation sufficient yet suitable features will be selected and derived in the first place. Typically when solving a time series problem it might be a good idea to identify the structure of the process, i.e., whether it follows the definition of ARMA(p, q). By this means it would enable us to directly generate features that theoretically take effect.

However, for a prediction problem it is not always necessary to do so because simply adding lags into the model delivers similar, yet sometimes even more effective performance. This improvement results from relaxed assumptions and more flexible determination of parameters. Here we will abandon the academic way and take a more empirical perspective. Also we will further talk about the determination of features closer in the following section.

The main steps include the removal of outliers, working with shops/items/cats objects and features, creating matrix as product of item/shop pairs within each month in the train set, getting monthly sales for each item, shop pair in the train set and merging it to the matrix, clipping item_cnt_month by (0,20), appending test to the matrix.

Through these feature engineering steps, raw data is preprocessed into our training set and carries purer information.

## 3.2 Random Forest Regressor

Features that have been decided and calculated will be sent into a machine learning model. Several models are popular in the world within data science but they have diverse applications due to the slight differences in their nature. Models that perform the calculation faster are usually preferred, if they deliver close fitting accuracy. That sounds like the reproduction of Darwinism: XGBoost replaced random forest, which was later surpassed by LightGBM.

However, they still generate quite different predictions, taken into consideration their similar $R^2$. Some people may also consider applying the mechanism of neural network, but in our mind this method relies too much on the structure and the independence of data and is more suitable when dealing with natural science problems where the relationship between features and the explained variable tend to have a more explicit form. Another reason to overlook neural network model is the intensiveness of computation.

The model based on Random Forest Regressor without ensemble learning reach a score of 0.88920 experimentally.

## 3.3 Ensemble Learning

To utilize the advantage of different models a new technique called ensembling is introduced. One common method to ensemble is stacking, where for each model part of the data will be replaced by the prediction of other models. With this approach more complicated model can be determined without much sacrifice in computational power.

Also stacking is more robust compared with any sole model especially where outliers will be carefully treated and transformed so they won't considerably affect the variation.

Due to time and condition constraints, we finally use a simple method of ensemble learning. We averaged five results of RandomForestRegressor. In this way, we achieved RMSE of 0.88186.

## 3.4 Parameter Optimization

It's also noticeable, although theoretically minor, that by choosing appropriate parameters a model can deliver different predictions. It's surprising that the modification of parameters can produce results that vary a lot, yet the total fitting accuracy is close to each other. This implies some others tricks, which we will talk about later.

Table 1: A quick look into different models

| Model | RMSE | Feature Engineering Method | Regressor | Advantages | Disadvantages |
|---|---|---|---|---|---|
| Feature engineering, xgboost | 0.91 | removal of outliers; working with shops/items/cats objects and features; converting the raw sales data to monthly sales | XGBoost | Available for parallel processing; Robust for missing value | Time-consuming for second derivative computing |
| Nested LSTM | 1.02 | converting the raw sales data to monthly sales | Nested LSTM | Well-suited to learn from experience when there are very long time lags of unknownsize between important events | Lack of refined feature engineering; Less attention to feature characteristics |
| Our model | **0.88** | removal of outliers; working with shops/items/cats objects and features; getting monthly sales for each item | Random Forest | Insensitiveness to hyper-parameters; Simple implement; Available for parallel processing; Robust for missing value | Possible progress with more experiment |

## 4    Experiments

In this section we present related experiments with respect to the processing of the sales data and model selection.

Based on existing kernels a natural idea emerges that a lower error might be possible with an altered combination of features. It took us two days of vain work to do so. It turns out that existing set of features has contained quite complete information that the absense of any feature can bring about significant loss of information, whereas the introduction of a new one can cause overlap and thus increase the variation of our estimation.

Another problem we have encountered is that, although the program can report the feature importance, it is still unwise to drop any feature based on the chart just because it lies near the bottom. Shrinking the feature set can have consequential changes: previously significant terms may become nearly irrelevant, and vise versa. It occurs to us that existing features may be highly correlated with each other.

Provided with the features selected in the kernel *Feature engineering, xgboost* [1], we tentatively substitute xgboostregressor for random forest algorithm, resulting in a remarkable leap of performance. It shows that the RF model with unmodified parameters can already enable us to reach an RMSE of 0.89293, superior to any other single model we have tried including GradientBoostRegressor(0.92677) and xgboostregressor(0.90684). After adjusting parameters using GridSearchCV our grade reaches top 10 with an RMSE of .88920.

Then we consider model fusion like ensemble learning to achieve a better score by combining random forest, XGBoost and Gradientboost in a 3-fold stacking firstly to maintain the diversity of model, yet it turns out to be worse than any single one with the result of 0.99504. The relatively big gap of performance between these models may account for this failed attempt, and thus we replace it with 5 different random forest models, which finally leads us to reach rank 5 in the Kaggle leaderboard.

Beyond above we have also tried acceleration using CUDA. Certain models like random forest are not designed for parallelization but an open library can still be found on GitHub yet with few repos. Other models such as LightGBM themselves are fast enough and are not necessary to be accelerated. Still, we attempt to move the estimation on a nVIDIA platform to see if a better result can be generated. Disappointedly we have encountered two cases of power failure and our program has never made it.

One shock is that, random forest, a merely legacy, outperforms all other models and leads us up among top 5. Slow and less robust, people now turn to emerging models like LightGBM and other boosting ones. But this time we learn that classical models should not be necessarily overlooked and one should always try as many model as possible before settling down a conclusion.

# 5   Conclusion

The main idea of our model is the emphasis on the characteristics of features. Random forest regressor is a simple regressor that satisfied our requirements, giving feature importance automatically and reach a perfect result. Ensemble learning attributes to our model as well. In addition to our good performance, our model can be simply implemented and trained.

It is worth considering that simple and conventional methods may make sense with appropriate preprocessing of the raw data. Future works for us is to extend our method to more application and hopefully learn more relative methods.

Through taking part in this competition, we not only get the idea of the strengths and weaknesses of different time series predicting methods, but also have a more practical experience in feature engineering and parameter optimization. Hopefully this experience will be our starting point of our future work on machine learning.